# Image Segmentation with transformers: An Overview, Challenges and Future

**Deepjyoti Chetia**[*], Debasish Dutta[2], Sanjib Kr Kalita[3]
[1,2,3]Dept. of Computer Science, Gauhati University, Assam, India.
*Corresponding Author: **Deepjyoti Chetia**              Email:deepjyotichetia@gauhati.ac.in

**Abstract:**
Image segmentation, a key task in computer vision, has traditionally relied on convolutional neural networks (CNNs), yet these models struggle with capturing complex spatial dependencies, objects with varying scales, need for manually crafted architecture components and contextual information. This paper explores the shortcomings of CNN-based models and the shift towards transformer architectures -to overcome those limitations. This work reviews state-of-the-art transformer-based segmentation models, addressing segmentation-specific challenges and their solutions. The paper discusses current challenges in transformer-based segmentation and outlines promising future trends, such as lightweight architectures and enhanced data efficiency. This survey serves as a guide for understanding the impact of transformers in advancing segmentation capabilities and overcoming the limitations of traditional models.

## 1. Introduction

Image segmentation is a fundamental task in computer vision that separates boundaries, objects, regions, or regions of interest within an image. It divides the image into meaningful segments so that computers can perceive and understand visual information like humans. These meaningful segments of an image help computers further analyze it in different CV areas. Image segmentation has been serving as a crucial step in numerous tasks such as general objection recognition, medical imaging, autonomous driving, remote sensing, and satellite imagery. Traditionally image segmentation problem was solved using thresholding, edge-based, region-based, clustering-based, watershed algorithm, graph-based and contour-based segmentation methods[1]. These methods were easy to implement, and computationally efficient but faced several issues, which made them less effective in dealing with real-world and complex images. A few of the limitations were difficulty in capturing global contexts, sensitivity to noise and intensity variations, problem handling unclear or overlapping boundaries, and manual tuning[2]. With the advent of machine learning (ML) algorithms such as K-Means Clustering, Gaussian Mixture Model, Random Forest, Support Vector Machine, Conditional Random Fields, and Markov Random Fields; several limitations faced by previous statistical, mathematical and image processing techniques were solved. These models were very effective in feature extraction, binary or multi-class classification, images with overlapping colours etc. However, these algorithms struggled with high-dimensional or complex patterns, handcrafted feature selections, less adaptive to diverse datasets, and poor generalization. With the development of different neural networks (NN), the scenario of image segmentation evolved significantly. Especially NNs such as Convolutional Neural Networks(CNN) which perform effectively in images and image-related tasks. CNN models such as Fully Convolutional Networks (FCN)[3], UNet[4], DeepLab[5], and Pyramid Scene Parsing Network[6] contributed substantially while overcoming the limitations of ML models. Curation of large datasets such as ImageNet[7], PASCAL-VOC[8], COCO[9], Cityscape[10], Liver Tumor Segmentation[11] coupled with pre-trained models using transfer learning techniques and fine-tuning NNs led to further advancement of image segmentation field. CNNs are highly effective for image segmentation because they can automatically extract features, learn hierarchical representations, ensure translation invariance, and support end-to-end learning. This allows them to process raw pixel data directly, identifying simple to complex features through different network layers, recognizing objects regardless of position, and training seamlessly from input to segmented output[4][3]. Despite their success, these models struggle with the understanding of global context which can affect the segmentation of large connected objects[12], huge computational cost during training, especially in deeper networks[13], over-fitting in models with many parameters which leads to poor performance in testing data[14]. Recently, transformers[15] have revolutionised Natural Language Processing (NLP) by introducing a new architecture that outperforms and overcomes previous models like Recurrent Neural Networks (RNNs) and Long Short-Term Memory networks (LSTMs)[16]. Transformers introduced a novel self-attention mechanism and can process various tokens in parallel. Transformers can capture long-range dependencies more effectively than previous architectures. Attention mechanisms help in understanding contextual relationships. Transformers achieved strong performance on many NLP tasks. This resulted in using transformer architecture in computer vision tasks. Early methods [17] [18] use self-attention with CNN layers to augment CNN. Then few studies have used self-attention to replace CNN layers[19][20]. After that, two studies have significantly improved CV tasks. The first one is Vision Transformer (ViT)[21], which is a pure transformer architecture that divides images into patches and processes these patches as tokens. This allows the model of the relationship between patches to understand the image globally. This model achieved state-of-the-art performance on multiple benchmark image recognition datasets. Another architecture is a detection transformer (DETR)[22], which is an encoder-decoder architecture that applies a transformer for object detection tasks. DETR combines CNN backbone(e.g. ResNet) for feature extraction using self-attention and utilises learned positional encodings, and queries to produce prediction. Inspired by these works, many researchers adopted these transformer architectures for various vision tasks.

## 1.1. Contribution

This work introduces recent advancements in transformer-based image segmentation methods. This work defines the task, datasets, metrics, CNN-based methods, limitations in CNN-based methods and how these limitations are overcome by transformer-based methods. Additionally, the study discusses the challenges of current segmentation transformer models and their future directions.

## 2. Background
## 2.1. Problem Definition

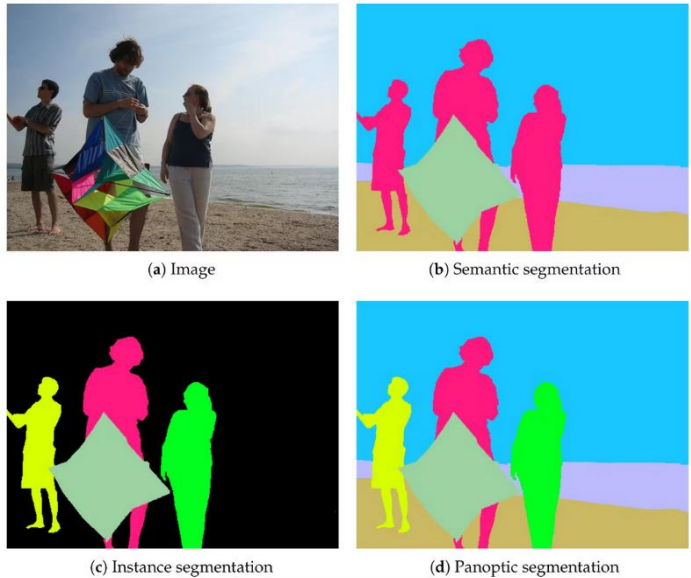

Figure 1: Types of image segmentation

Given an input image $\in R^{H \times W \times 3}$, the goal of image segmentation is to output a group of masks $\{y_i\}_{i=1}^{G} = \{(m_i, c_i)\}_{i=1}^{G}$ denotes the ground truth class label of the binary mask $m_i$ and G is the number of masks, H × W are the spatial size.

Image segmentation can be classified into three types: semantic segmentation(SS), instance segmentation(IS) and panoptic segmentation(PS). SS assigns a label to each pixel in an image so that pixels belonging to the same object category are grouped. Multiple instances of the same objects are treated as a single entity in SS. Every pixel is labelled according to its object class. E.g. all pixels representing "people" are marked with the same label. IS identifies and detects the boundary of each object in an image. IS divides each instance of the image even if the instances belong to the same class. IS models usually require more computation than SS. PS utilizes both semantic and instance segmentation approaches. PS gives a complete view of the image by assigning each class label(SS) and dividing individual instances of objects (PS). This unified representation ensures that every pixel is assigned to a class whether as a part of the object or background. Figure 1 shows the differences among type of image segmentation.

## 2.2. Challenges in Image Segmentation

**Dataset annotation**: For large-size datasets creating pixel-level annotation is time-consuming, labour-intensive and expensive. Which in turn leads to poor performance in model evaluation. Also, in some domains such as medical imaging, annotating requires domain experts for ground truth.

**Difficulties with small or thin objects**: Detection and segmentation of thin objects such as blood vessels, and hair requires models to preserve finer details. Which is a big challenge for segmentation architectures.

**Boundary Accuracy**: Tasks such as instance segmentation require precise boundary detection for different objects in an image. Identifying the boundary for objects with irregular shapes and ambiguous edges becomes a challenge for segmentation methods.

**Model generalization Across Domains**: Neural networks trained on one dataset often fail to generalize well to real-world scenarios. Deploying such models results in poor performance.

## 2.3. Datasets

The curation of different datasets has proven beneficial for addressing the problem of image segmentation. Image segmentation tasks require high-quality, annotated datasets to train and evaluate the models effectively. A brief comparison of some of the popular datasets is given in Table 1.

Table 1  Benchmarking datasets for Image segmentation

| Dataset | Release | Paradigm | Description | Size | Applications | Challenges |
|---|---|---|---|---|---|---|
| COCO | 2014 | PS, SS | Complex scenes with multiple objects. | 200,000+ images, 1.5M+ object instances | General-purpose segmentation | High diversity, complex backgrounds. |
| PASCAL VOC | 2007 | SS, IS | Benchmark dataset for basic segmentation tasks. | ~20,000 images, 20 classes | Object detection, segmentation | A limited number of classes |
| Cityscapes | 2016 | SS, IS | Urban street scenes captured from vehicles. | 5,000 finely, 20,000 coarsely annotated | Autonomous driving, smart cities | High-resolution images |
| ADE20K [23] | 2017 | SS | A diverse range of scenes and objects. | 25,000+ images, 150 classes | Scene parsing, diverse contexts | Requires models to generalize across a wide context. |
| BRATS [24] | 2012 | Medical Image | Brain tumour segmentation using multi-modal MRI scans. | 300+ cases, multi-modal MRI | Brain tumour diagnosis, research | Complex tumour structures |

## 2.4. Evaluation

To evaluate the performance of a segmentation model specific metrics are required. These metrics can accurately reflect how well the model is segmenting different parts of an image. Pixel Accuracy(PA) measures the ratio of correctly classified pixels to the total pixels, providing an overall accuracy metric. Mean Pixel Accuracy (mPA) averages the class-wise accuracy, addressing class imbalance by equally weighting each class. Intersection over Union (IoU) calculates the overlap between the predicted and ground truth regions for a class, divided by their union. Mean IoU (mIoU) averages the IoU across all classes, offering a holistic evaluation. The Dice Coefficient is a similarity metric, often used in medical imaging, that emphasizes the overlap by weighing it twice as much as the union. Precision assesses the proportion of correctly predicted positive pixels, while Recall quantifies how well the model identifies all relevant pixels. The F1 Score, the harmonic mean of precision and recall, balances these two metrics to provide a single performance measure, especially useful in imbalanced datasets.

## 2.5. Traditional Methods

Before the introduction of transformers in segmentation, various statistical, machine learning, and deep learning methods were utilized. Over time, these methods evolved, significantly improving segmentation efficiency, but at the expense of increased computational cost, particularly for deep learning approaches. Neural Networks such as CNN, and RNN were used in many CV tasks but CNN has outperformed RNN [25]. Table 2 gives an overview of methods used in image segmentation before transformers were used.

Table 2 Traditional image segmentation models

| Method | Year | Paradigm | Metrics Used | Results | Dataset | Application |
|---|---|---|---|---|---|---|
| Fully Convolutional Networks (FCN) | 2014 | Semantic | mIoU | mIoU of 62.2% | PASCAL VOC | General-purpose segmentation |
| U-Net | 2015 | Semantic | Dice Coefficient, Jaccard Index | Dice coefficient > 0.8 | ISBI | Medical imaging, bioinformatics |
| DeepLab (v1-v3) | 2015-18 | Semantic | mIoU, Pixel Accuracy | mIoU of 84.56% | Cityscapes | Autonomous driving, satellite imagery |
| SegNet [26] | 2015 | Semantic | Pixel Accuracy, | mIoU ~74% | CamVid | Robotics, |

| Method | Year | Paradigm | Metrics Used | Results | Dataset | Application |
|---|---|---|---|---|---|---|
| | | | mIoU | | | autonomous driving |
| Mask R-CNN [he2017] | 2017 | Instance | (AP), mAP, mIoU | AP of 37.1% and mIoU ~50% | COCO | Autonomous driving, video analysis |
| Pyramid Scene Parsing Network (PSPNet) [6] | 2017 | Semantic | mIoU, Global Pixel Accuracy | mIoU of 85.4% | ADE20K | Scene understanding, autonomous vehicles |
| DenseNet-based Models [27] | 2017 | Semantic | mIoU, Dice Coefficient | Dice coefficient ~0.9 | medical imaging | Medical imaging, remote sensing |
| HRNet [28] | 2019 | Semantic | mIoU | mIoU of 81.6% | Cityscapes | Human pose estimation, scene parsing |

## 3. Trasformer Network

The transformer introduced in the paper [15] revolutionized NLP tasks and utilizes an attention mechanism, which enables the model to process and weigh the importance of all input tokens relative to each other, allowing the model for long-range dependency learning. Later, transformers were employed in CV tasks such as classification, detection and segmentation to outperform existing architectures [21]. A transformer which is essentially an encoder-decoder architecture, at the core of transformer is the self-attention, which computes the relationship between input elements(token) to understand interdependence. Each input is transformed into Query(Q), Key(K), and Value vectors, which are used to calculate attention scores.

$$Attention(Q, K, V) = softmax\left(\frac{QK^T}{\sqrt{d_k}}\right)$$

Where $d_k$ is the dimension of the Key vector, used for scaling to stabilize gradients. To enhance the model's ability to focus on different parts of the input, transformers employ multiple attention "heads," where each head operates on different projections of Q, K, and V. The outputs are concatenated and linearly transformed. Unlike previous sequential models such as RNNs, transformers process inputs in a parallel manner. So, to maintain the order positional encodings are added to input embeddings. Each transformer layer includes a feed-forward network(FFN with a non-linear activation function (ReLU) for feature extraction and non-linear transformation. To prevent exploding or vanishing gradient and training efficiency, normalization is applied. Skip connections around attention and FFN layers enable the flow of information directly to deeper layers.

### 3.1. Transformers for Image Segmentation

Transformers have been adapted for image segmentation through different architectural adjustments, utilizing their self-attention mechanism to address the dense prediction needs for segmentation tasks. Transformers originally designed for sequences, process images by dividing the image as patches[21]. Each patch is flattened and linearly projected to create embeddings. These embedding coupled with positional encodings, fed into the transformer encoder. This approach is used in models like ViT[21] and Segmenter[29]. Some architectures such as Fully Transformer Network [30] and Pyramid Vision Transformer[31] added pyramid structure that captures both fine details and global context. Hierarchical representations help adapt transformers to dense tasks like segmentation. In decoders like Mask2Former, adding mechanisms such as masked attention, helped generate accurate segmentation masks [32]. Mask2Former outperformed the best-specialized models in every segmentation type (SS, IS, PS). These architectural adjustments enabled models for effective image segmentation.

### 3.2. State-of-the-art transformer models

In this section, a popular SOTA image segmentation model based on transformer architecture is given in Table 3 with their results and advantages.

*Table 3 SOTA transformer image segmentation models*

| Model | Year | Key Features | Dataset & Results | Advantages |
|---|---|---|---|---|
| Vision Transformer (ViT) [21] | 2020 | Divides image into patches and applies transformer encoder for feature extraction. | mIoU ~72% on ADE20K. | Strong global feature representation but requires fine-tuning for dense tasks like segmentation. |
| SETR (Segmenter) [33] | 2021 | Uses a pure transformer encoder for pixel-level prediction; avoids convolution layers. | mIoU ~79.4% on Cityscapes. | Captures global context better than CNNs. |
| DETR (Detection Transformer) [22] | 2020 | Iinitially designed for object detection; extended for instance segmentation using bipartite matching. | mIoU ~44% on the COCO dataset. | Integrates detection and segmentation seamlessly. |
| Swin Transformer [34] | 2021 | Introduces hierarchical representations and shifted windows for efficient computation. | mIoU ~84.1% on ADE20K. | Combines efficiency with powerful feature extraction for segmentation tasks. |
| Segmenter [29] | 2021 | Leverages ViT-based architecture with a segmentation head. | mIoU ~80% on Pascal VOC. | Simplifies segmentation pipeline with transformer-based backbones. |
| Mask2Former [32] | 2022 | Extends MaskFormer with enhanced dynamic attention mechanisms for better mask prediction. | AP of ~46.7% on COCO. | Excels in an instance, semantic, and panoptic segmentation tasks. |

### 3.3. Overcoming the challenges of traditional Models

Transformers have made significant improvements in segmentation by overcoming limitations in traditional approaches, especially in CNN-based architectures. CNNs use convolutional filters to learn features, these filters have a limited receptive field, which grows slowly with the network depth. This makes it challenging to capture long-range dependencies between distant objects or parts of an image. Transformers with self-attention capture global dependencies across an entire image. Each patch or pixel can relate to every other, allowing the model to capture context from across the image. Since CNNs use fixed-size filters, they struggle with objects with various scales. Transformer-based methods such as Swin[34] and SegFormer [35] introduced a multi-scaled hierarchical structure which allows the model to learn relationships across scales. So, the models can handle small or large objects effectively. Previous methods rely on anchor boxes, region proposals, fixed-feature pyramids in object detection and instance segmentation. This hinders model generalizability since these components need careful manual tuning. Transformer-based architectures, such as DETR [22], remove the need for traditional components like region proposal networks, and anchor boxes. Instead, DETR predicts object locations and segmentation masks directly in a single, unified pipeline. This end-to-end framework reduces architectural complexity, simplifies training and inference processes, and increases model generalisation and adaptability.

## 4. Challenges and Future Trends

Despite the success of transformers in image segmentation, there lie many challenges in architecture such as large data dependency, high computational cost and difficulty in generalizing to small targets. Transformers also have poor model interpretability, making it difficult in critical domains such as healthcare. These disadvantages create the need for more efficient models that address computational cost and memory constraints and adapt to small datasets. Combining traditional methods with transformers may be a future avenue. By effectively addressing the current challenges and embracing emerging trends, transformer-based image segmentation is set to make a transformative impact on the advancement of the domain.